\documentclass[10pt,twocolumn,letterpaper]{article}

\usepackage[pagenumbers]{cvpr} 

\usepackage[dvipsnames]{xcolor}

\definecolor{cvprblue}{rgb}{0.21,0.49,0.74}
\usepackage[pagebackref,breaklinks,colorlinks,citecolor=cvprblue]{hyperref}
\usepackage{dsfont, amsfonts}
\usepackage{booktabs, multirow}
\usepackage{marvosym, hyperref}
\def\paperID{14915} 
\def\confName{CVPR}
\def\confYear{2024}

\title{Compact 3D Gaussian Representation for Radiance Field}

\author{Joo Chan Lee$^{1}$\quad Daniel Rho$^{2}$\quad Xiangyu Sun$^{1}$\quad Jong Hwan Ko$^{1}$\textsuperscript{\Letter}\quad Eunbyung Park$^{1}$\textsuperscript{\Letter}\\\\
Sungkyunkwan University$^{1}$, KT$^{2}$
}

\begin{document}

\twocolumn[{%
\renewcommand\twocolumn[1][]{#1}%
\maketitle
\begin{center}
    \centering
    \captionsetup{type=figure}
    \vspace{-1em}
    \includegraphics[width=1.0\linewidth]{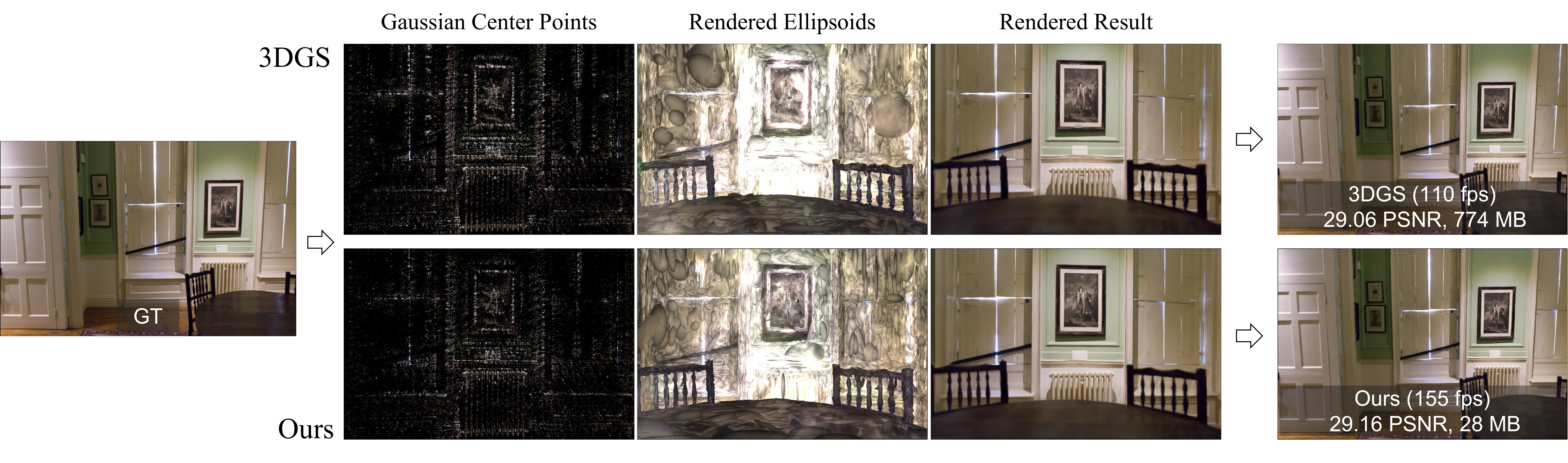}
    \vspace{-2em}
    \captionof{figure}{Our method achieves reduced storage and faster rendering speed while maintaining high-quality reconstruction of 3DGS~\cite{3dgs}. The core idea is to effectively remove the redundant Gaussians that do not significantly contribute to the overall performance (the sparser distribution of Gaussian points and reduced ellipsoid redundancy shown in the figure). We also introduce a more compact representation of Gaussian attributes, resulting in markedly improved storage efficiency and rendering speed.}
    \label{fig:demo}
\end{center}%
}]
\begin{abstract}
Neural Radiance Fields (NeRFs) have demonstrated remarkable potential in capturing complex 3D scenes with high fidelity. However, one persistent challenge that hinders the widespread adoption of NeRFs is the computational bottleneck due to the volumetric rendering. On the other hand, 3D Gaussian splatting (3DGS) has recently emerged as an alternative representation that leverages a 3D Gaussisan-based representation and adopts the rasterization pipeline to render the images rather than volumetric rendering, achieving very fast rendering speed and promising image quality.
However, a significant drawback arises as 3DGS entails a substantial number of 3D Gaussians to maintain the high fidelity of the rendered images, which requires a large amount of memory and storage.
To address this critical issue, we place a specific emphasis on two key objectives: reducing the number of Gaussian points without sacrificing performance and compressing the Gaussian attributes, such as view-dependent color and covariance.
To this end, we propose a learnable mask strategy that significantly reduces the number of Gaussians while preserving high performance. In addition, we propose a compact but effective representation of view-dependent color by employing a grid-based neural field rather than relying on spherical harmonics. Finally, we learn codebooks to compactly represent the geometric attributes of Gaussian by vector quantization.
With model compression techniques such as quantization and entropy coding, we consistently show over 25$\times$ reduced storage and enhanced rendering speed, while maintaining the quality of the scene representation, compared to 3DGS.
Our work provides a comprehensive framework for 3D scene representation, achieving high performance, fast training, compactness, and real-time rendering.
Our project page is available at \href{https://maincold2.github.io/c3dgs/}{https://maincold2.github.io/c3dgs/}.
\end{abstract}

\section{Introduction}
The field of neural rendering has witnessed substantial advancements in recent years, driven by the pursuit of rendering photorealistic 3D scenes from limited input data. Among the pioneering approaches, Neural Radiance Field (NeRF)~\cite{nerf} has gained considerable attention for its remarkable ability to generate high-fidelity images and 3D reconstructions of scenes from only a collection of 2D images in diverse applications. Follow-up research efforts have been dedicated to improving image quality~\cite{mip-nerf, instant-ngp}, accelerating training and rendering speed~\cite{killonerf, fridovich2022plenoxels, tensorf, kplanes, instant-ngp, dsnerf}, and reducing memory and storage footprints~\cite{masked, VQAD}.

Despite the massive efforts, one persistent challenge that hinders the widespread adoption of NeRFs is the computational bottleneck due to the volumetric rendering. 
Since it demands dense point sampling along the ray to render a pixel, which requires significant computational resources, NeRFs often fail to achieve real-time rendering on hand-held devices or low-end GPUs.
This challenge limits their use in practical scenarios where fast rendering speed is essential, such as various interactive 3D applications.

3D Gaussian splatting (3DGS)~\cite{3dgs} has recently emerged as an alternative representation that achieved very fast rendering speed and promising image quality.
This approach leverages a point-based representation associated with 3D Gaussian attributes and adopts the rasterization pipeline to render the images rather than volumetric rendering.
Highly optimized customized cuda kernels to maximize the parallelism and clever algorithmic tricks enable unprecedented rendering speed without compromising the image quality.
However, a significant drawback arises as 3DGS entails a substantial number of 3D Gaussians to maintain the high-fidelity of the rendered images (\cref{fig:demo}), which requires a large amount of memory and storage (e.g., often $>1$GB for representing a large real-world scene).


To address the critical large memory and storage issue in 3DGS, we propose a compact 3D Gaussian representation framework.
This approach significantly improves memory and storage efficiency while showing high-quality reconstruction, fast training speed, and real-time rendering, as shown in ~\cref{fig:demo}.
We place a specific emphasis on two key objectives.
First, we aim to reduce the number of Gaussian points required for scene representation without sacrificing performance.
The number of Gaussians increases with regular densification processes consisting of cloning and splitting Gaussians, and it was a crucial component in representing the fine details of scenes.
However, we observed that the current densification algorithm produces many redundant and unnecessary Gaussians, resulting in high memory and storage requirements.
We introduce a novel volume-based masking strategy that identifies and removes non-essential Gaussians that have minimal impact on overall performance.
With the proposed masking method, we learn to reduce the number of Gaussians while achieving high performance during training.
In addition to the efficient memory and storage usage, we can achieve faster rendering speed since the computational complexity is linearly proportional to the number of Gaussians.

Second, we propose compressing the Gaussian attributes, such as view-dependent color and covariance. 
In the original 3DGS, each Gaussian has its own attributes, and it does not exploit spatial redundancy, which has been widely utilized for various types of signal compression.
For example, neighboring Gaussians may share similar color attributes, and we can reuse similar colors from neighboring Gaussians.
Given this motivation, we incorporate a grid-based neural field to efficiently represent view-dependent colors rather than using per Gaussian color attributes.
When provided with the query Gaussian points, we extract the color attribute from the compact grid representation, avoiding the need to store it for each Gaussian separately.
For our initial approach, we opt for a hash-based grid representation (Instant NGP~\cite{instant-ngp}) from among several candidates due to its compactness and fast processing speed. This choice has led to a significant reduction in the spatial complexity of 3DGS.

In contrast to the color attribute, the majority of Gaussians exhibit similar geometry, with limited variation in scale and rotation attributes.
3DGS represents a scene with numerous small Gaussians collectively, and each Gaussian primitive is not expected to show high diversity.
Therefore, we introduce a codebook-based approach for modeling the geometry of Gaussians.
It learns to find similar patterns or geometry shared across each scene and only stores the codebook index for each Gaussian, resulting in a very compact representation.
Moreover, because the codebook size can be quite small, the spatial and computational overhead during training are not significant.


We have extensively tested our proposed compact 3D Gaussian representation on various datasets including real and synthetic scenes. Throughout the experiments regardless of dataset, we consistently showed about 15$\times$ reduced storage and enhanced rendering speed, while maintaining the quality of the scene representation, compared to 3DGS.
Furthermore, our method can benefit from simple post-processings such as quantization and entropy coding, consequently achieving over 25$\times$ compression throughout the datasets.
Especially for the evaluation on Deep Blending~\cite{db}, a real-world dataset, we outperform 3DGS in terms of the reconstruction quality (measured in PSNR), notably with over 28$\times$ enhancement in storage efficiency and nearly 40\% increase in rendering speed, setting a new state-of-the-art benchmark.

\section{Related Work}
\subsection{Neural Radiance Fields}
Neural radiance fields (NeRFs) have significantly expanded the horizons of 3D scene reconstruction and novel view synthesis. NeRF~\cite{nerf} introduced a novel approach to synthesizing novel views of 3D scenes, representing volume features by utilizing Multilayer Perceptrons (MLPs) and introducing volumetric rendering. Since its inception, various works have been proposed to enhance performance in diverse scenarios, such as different resolutions of reconstruction~\cite{mip-nerf, mip360}, the reduced number of training samples~\cite{yu2021pixelnerf, niemeyer2022regnerf, sinnerf, ibrnet, depthprior}, and reconstruction of large realistic scenes~\cite{rawnerf, blocknerf} and dynamic scenes~\cite{dnerf,Li_2021_CVPR, Li_2022_CVPR, Gao_2021_ICCV}. However, NeRF's reliance on MLP has been a bottleneck, particularly causing slow training and inference.

In an effort to address the limitations, grid-based methods emerged as a promising alternative. These approaches using explicit voxel grid structures~\cite{fridovich2022plenoxels, DVGO, zipnerf, tineuvox, nsvf, devrf} have demonstrated a significant improvement in training speed compared to traditional MLP-based NeRF methods. Nevertheless, despite this advancement, grid-based methods still suffer from relatively slow inference speeds and, more importantly, require large amounts of memory. This has been a substantial hurdle in advancing towards more practical and widely applicable solutions.

Subsequent research efforts have been directed toward the reduction of the memory footprint while maintaining or even enhancing the performance quality by grid factorization~\cite{tensorf, EG3D, kplanes, multensorf, strivec, mipgrid}, hash grids~\cite{instant-ngp}, grid quantization~\cite{VQAD, birf} or pruning~\cite{fridovich2022plenoxels, masked}. These methods have also been instrumental in the fast training of 3D scene representation, thereby making more efficient use of computational resources. However, a persistent challenge that remains is the ability to achieve real-time rendering of large-scale scenes. The volumetric sampling inherent in these methods, despite their advancements, still poses a limitation.

\subsection{Point-based Rendering and Radiance Field}
Point-NeRF~\cite{pointnerf} employed points to represent a radiance field with volumetric rendering. Although it shows promising performance, the volume rendering hinders the real-time rendering.
NeRF-style volumetric rendering and point-based $\alpha$-blending fundamentally share the same model for rendering images but differ significantly in their rendering algorithms~\cite{3dgs}. 
NeRFs offer a continuous feature representation of the entire volume as empty or occupied spaces, which necessitate costly volumetric sampling to render a pixel, leading to high computational demands.
In contrast, points provide an unstructured, discrete representation of a volume geometry by the creation, destruction, and movement of points, and a pixel is rendered by blending several ordered points overlapping the pixel.
This is achieved by optimizing opacity and positions~\cite{kopanas2021}, thus bypassing the limitations inherent in full volumetric representations and achieving remarkably fast rendering.

Point-based methods have been widely used in rendering 3D scenes, where the simplest form is point clouds. 
However, point clouds can lead to visual artifacts such as holes and aliasing.
To mitigate this, point-based neural rendering methods have been proposed, processing the points through rasterization-based point splatting and differentiable rasterization~\cite{yifan, wiles, pulsar}.
The points were represented by neural features and rendered with CNNs~\cite{aliev, kopanas2021, meshry}.
However, these methods heavily rely on Multi-View Stereo (MVS) for initial geometry, inheriting its limitations, especially in challenging scenarios like areas lacking features, shiny surfaces, or fine structures.

Neural Point Catacaustics~\cite{kopanas2022} addressed the issue of view-dependent effect through the use of an MLP, yet it still depends on MVS geometry for its input. 
Without the need for MVS, \citet{zhang2022} incorporated Spherical Harmonics (SH) for directional control. However, this method is constrained to managing scenes with only a single object and requires the use of masks during its initialization phase. 
Recently, 3D Gaussian Splatting (3DGS)~\cite{3dgs} proposed using 3D Gaussians as primitives for real-time neural rendering.
3DGS utilized highly optimized custom CUDA kernels and ingenious algorithmic approaches, it achieves unparalleled rendering speed without sacrificing image quality.

\begin{figure*}[t]
    \begin{center}
    \includegraphics[width=1.0\linewidth]{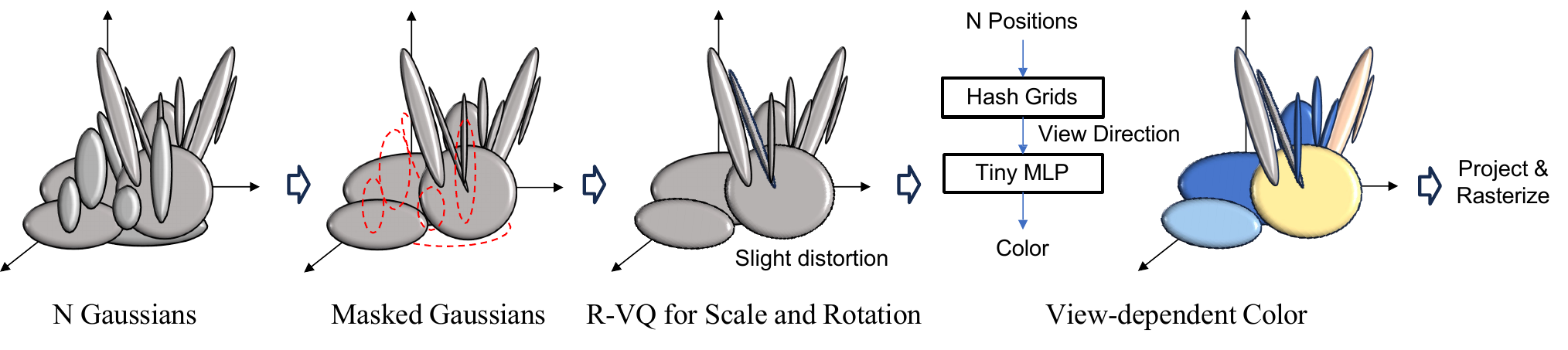}
    \end{center}
    \vspace{-1.5em}
    \caption{The detailed architecture of our proposed compact 3D Gaussian.}
    \vspace{-1.5em}
\label{fig:arch}
\end{figure*}

While 3DGS does not require dense sampling for each ray, it does require a substantial number of 3D Gaussians to maintain a high level of quality in the resulting rendered images.
Additionally, since each Gaussian consists of several rendering-related attributes like covariance matrices and SH with high degrees, 3DGS demands significant memory and storage resources, e.g., exceeding 1GB for a realistic scene.
Our work aims to alleviate this parameter-intensive requirement while preserving high rendering quality, fast training, and real-time rendering.

\section{Method}
\noindent\textbf{Background.}
In our approach, we build upon the foundation of 3D Gaussian Splatting (3DGS)~\cite{3dgs}, a point-based representation associated with 3D Gaussian attributes for representing 3D scenes.
Each Gaussian represents 3D position, opacity, geometry (3D scale and 3D rotation represented as a quaternion), and spherical harmonics (SH) for view-dependent color.
3DGS constructs initial 3D Gaussians derived from the sparse data points obtained by Structure-from-Motion (SfM), such as COLMAP~\cite{colmap}. 
These Gaussians are cloned, split, pruned, and refined towards enhancing the anisotropic covariance for a precise depiction of the scene.
This training process is based on the gradients from the differentiable rendering without unnecessary computation in empty space, which accelerates training and rendering.
However, 3DGS's high-quality reconstruction comes at the cost of memory and storage requirements, particularly with numerous Gaussians increased during training and their associated attributes.

\noindent\textbf{Overall architecture.}
Our primary objectives are to 1) reduce the number of Gaussians and 2) represent attributes compactly while retaining the original performance.
To this end, along with the optimization process, we mask out Gaussians that minimally impact performance and represent geometric attributes by codebooks, as shown in \cref{fig:arch}.
We represent the color attributes using a grid-based neural field rather than storing them directly per each Gaussian.
For geometry attributes, such as scale and rotation, we propose using a codebook-based method that can fully exploit the limited variations of these attributes.
Finally, a small number of Gaussians with compact attributes are then used for the subsequent rendering steps, including projection and rasterization to render images.

\begin{figure}[t]
    \begin{center}
    \includegraphics[width=1.0\linewidth]{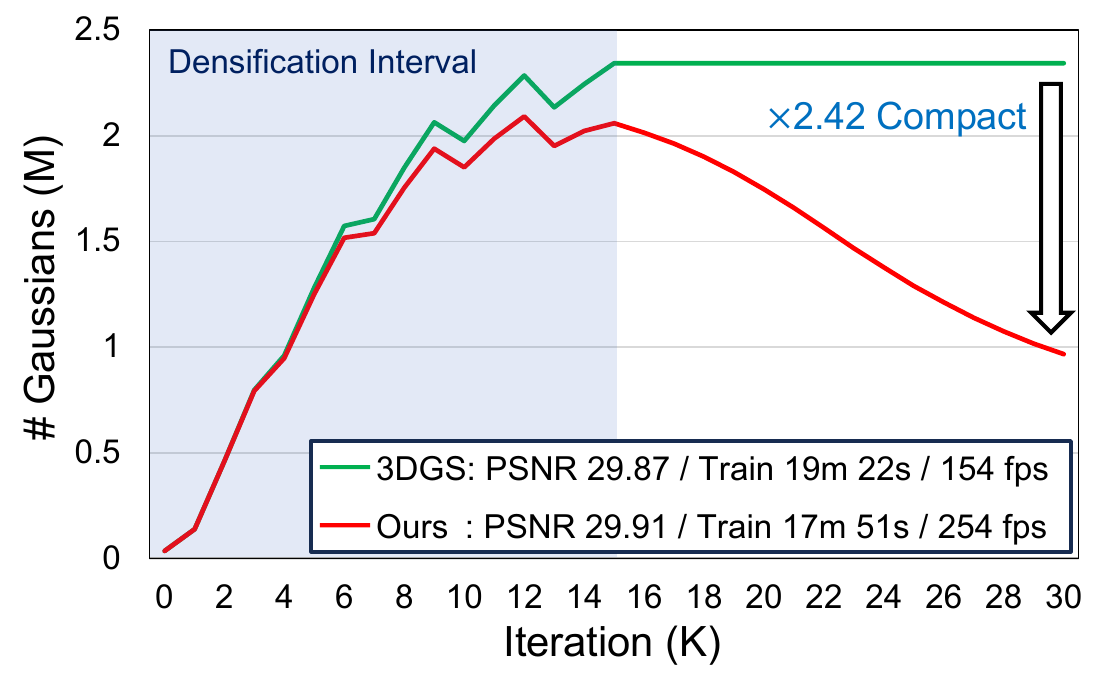}
    \end{center}
    \vspace{-2em}
    \caption{Visualization of the varying count of Gaussians during training (\textit{Bonsai} scene). `\# Gaussians' denotes the number of Gaussians.}
    \vspace{-1.5em}
\label{fig:mask}
\end{figure}

\subsection{Gaussian Volume Mask}
3DGS originally densifies Gaussians with large gradients by cloning or splitting.
To regulate the increase in the number of Gaussians, opacities are set to a small number at every specific interval, and after some iterations, those with still minimal opacities are removed.
Although this opacity-based control effectively eliminates some floaters, we empirically found that redundant Gaussians still exist significantly ($\times$2.42 Gaussians show similar performance in \cref{fig:mask}).

The scale attribute of each Gaussian determines its 3D volume, which is then reflected in the rendering process. 
Small-sized Gaussians, due to their minimal volume, have a negligible contribution to the overall rendering quality, often to the point where their effect is essentially imperceptible. 
In such cases, it becomes highly beneficial to identify and remove such unessential Gaussians.

To this end, we propose a learnable masking of Gaussians based on their volume as well as transparency.
We apply binary masks not only on the opacities $o\in [0,1]^{N}$ but also on the non-negative scale attributes $s\in \mathbb{R}_{+}^{N\times3}$ that determine the volume geometry of $N$ Gaussians, where $N$ may vary with densification of Gaussians.
We introduce an additional mask parameter $m\in \mathbb{R}^{N}$, based on which we generate binary masks $M\in \{0,1\}^{N}$.
As it is not feasible to calculate gradients from binarized masks, we employ the straight-through estimator~\cite{ste}.
More specifically, the masked scale $\hat{s} \in \mathbb{R}_{+}^{N\times3}$ and the masked opacity $\hat{o} \in [0,1]^N$ are formulated as follows,
\begin{align}
    &M_n = \operatorname*{sg}(\mathds{1}[\sigma(m_n) > \epsilon] - \sigma(m_n)) + \sigma(m_n),  \\
    &\hat{s}_n = M_n s_n,\quad \hat{o}_n = M_n o_n,
\end{align}
where $n$ is the index of the Gaussian, $\epsilon$ is the masking threshold, $\operatorname*{sg}(\cdot)$ is the stop gradient operator, and $\mathds{1}[\cdot]$ and $\sigma(\cdot)$ are indicator and sigmoid function, respectively.
This method allows for the incorporation of masking effects based on Gaussian volume and transparency in rendering. Considering both aspects together leads to more effective masking compared to considering either aspect alone.

We balance the accurate rendering and the number of Gaussians eliminated during training by adding masking loss $L_m$ as follows,
\begin{align}
    &L_m = {1\over N} \sum_{n=1}^{N}\sigma({m_n}).
\end{align}

At every densification, we eliminate the Gaussians according to the binary mask.
Furthermore, unlike the original 3DGS that stops densifying in the middle of the training and retains the number of Gaussians to the end, we consistently mask out along with the entire training process, reducing unessential Gaussians effectively and ensuring efficient computation with low GPU memory throughout the training phase (\cref{fig:mask}).
Once training is completed, the mask parameter $m$ does not need to be stored since we removed the masked Gaussians.

\begin{figure*}[t]
    \begin{center}
    \includegraphics[width=1.0\linewidth]{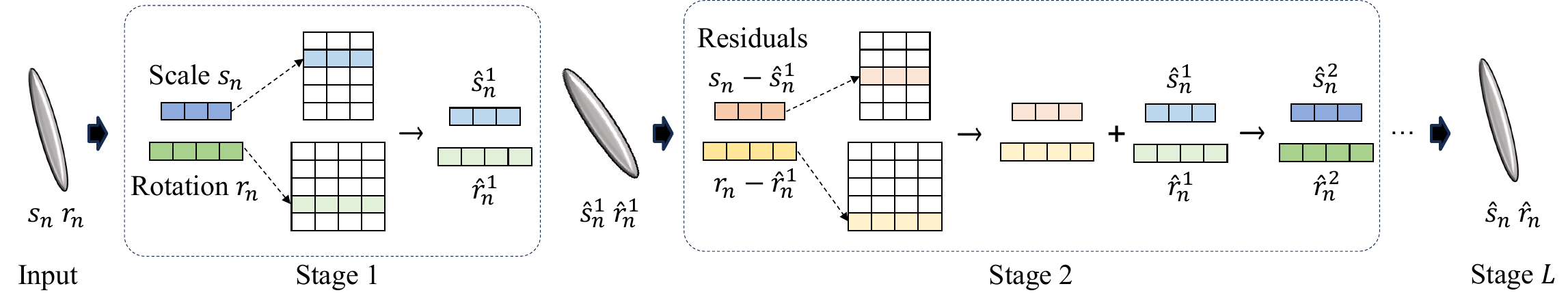}
    \end{center}
    \vspace{-2em}
    \caption{The detailed process of R-VQ to represent the scale and rotation of Gaussians. In the first stage, the scale and rotation vectors are compared to codes in each codebook, with the closest code identified as the result. In the next stage, the residual between the original vector and the first stage's result is compared with another codebook. This process is repeated up to the final stage, as a result, the selected indices and the codebook from each stage collectively represent the original vector.}
    \vspace{-1em}
\label{fig:rvq}
\end{figure*}

\subsection{Geometry Codebook}
A number of Gaussians collectively construct a single scene, where similar geometric components can be shared throughout the entire volume.
We have observed that the geometrical shapes of most Gaussians are very similar, showing only minor differences in scale and rotation characteristics. 
In addition, a scene is composed of many small Gaussians, and each Gaussian primitive is not expected to exhibit a wide range of diversity.
Given this motivation, we propose a codebook learned to represent representative geometric attributes, including scale and rotation, by employing vector quantization (VQ)~\cite{vq}.
As naively applying vector quantization requires computational complexity and large GPU memory~\cite{soundstream}, we adopt residual vector quantization (R-VQ)~\cite{soundstream} that cascades $L$ stages of VQ with codebook size $C$ (\cref{fig:rvq}), formulated as follows,
\begin{align}
    &\hat{r}_n^{l} = \sum_{k=1}^{l}\mathcal{Z}^k[i^k],\,\,\,\, l \in \{1,...,L\}, \\
    &i_n^l = \operatorname*{argmin}_{k} ||\mathcal{Z}^{l}[{k}]-(r_n - \hat{r}_n^{l-1})||_2^2,\,\,\,\, \hat{r}_n^0 = \vec{0}
\end{align}
where $r \in\mathbb{R}^{N\times4}$ is the input rotation vector, $\hat{r}^{l} \in\mathbb{R}^{N\times4}$ is the output rotation vector after $l$ quantization stages, and $n$ is the index of the Gaussian. $\mathcal{Z}^l\in\mathbb{R}^{C \times 4}$ is the codebook at the stage $l$, $i^l \in \{0,…,C-1\}^{N}$ is the selected indices of the codebook at the stage $l$, and $\mathcal{Z}[i] \in \mathbb{R}^4$ represents the vector at index $i$ of the codebook $\mathcal{Z}$.

The objective function for training the codebooks is as follows,
\begin{align}
    &L_r = {1 \over NC} \sum_{k=1}^{L} \sum_{n=1}^{N} ||\operatorname*{sg}[r_n -\hat{r}_n^{k-1}] - \mathcal{Z}^k[i_n^k]||_2^2,
\end{align}
where $\operatorname*{sg}[\cdot]$ is the stop-gradient operator.
We use the output from the final stage $\hat{r}^{L}$ (we will omit the superscript $L$ for brevity from now onwards), and the R-VQ process is similarly applied to scale $s$ before masking (we also similarly use the objective function for scale $L_s$).

\subsection{Compact View-dependent Color}
Each Gaussian in 3DGS requires 48 of the total 59 parameters to represent SH (max 3 degrees) to model the different colors according to the viewing direction.
Instead of using the naive and parameter-inefficient approach, we propose representing the view-dependent color of each Gaussian by exploiting a grid-based neural field.
To this end, we contract the unbounded positions $p\in\mathbb{R}^{N\times 3}$ to the bounded range, motivated by mip-NeRF 360~\cite{mip360}, and compute the 3D view direction $d\in\mathbb{R}^{3}$ for each Gaussian based on the camera center point.
We exploit hash grids~\cite{instant-ngp} followed by a tiny MLP to represent color.
Here, we input positions into the hash grids, and then the resulting feature and the view direction are fed into the MLP.
More formally, view-dependent color $c_n(\cdot)$ of Gaussian at position $p_n \in\mathbb{R}^3$ can be expressed as,
\begin{align}
    &c_n(d) = f(\operatorname*{contract}(p_n), d; \theta),
\end{align}\vspace{-2em}
\begin{align}
    \operatorname*{contract}(p_n) = \begin{cases} \,\, p_n & ||p_n||\le 1 \\
    \left(2-{1\over||p_n||}\right) \left({p_n\over||p_n||}\right) & ||p_n||>1,
\end{cases}
\end{align}
where $f(\cdot;\theta), \operatorname*{contract}(\cdot):\mathbb{R}^3\rightarrow\mathbb{R}^3$ stand for the neural field with parameters $\theta$, and the contraction function, respectively.
We use the 0-degree components of SH (the same number of channels as RGB, but not view-dependent) and then convert them into RGB colors due to the slightly increased performance compared to representing the RGB color directly.

\begin{table*}[ht]
\centering
\caption{Qualitative results of the proposed method evaluated on Mip-NeRF 360 and Tanks\&Temples datasets. We reported the numbers of baselines from the original paper (denoted as 3DGS), which were run on an NVIDIA A6000 GPU. For a fair comparison, we re-evaluate 3DGS with the same training configurations as our method using an NVIDIA A100 GPU (denoted as 3DGS*).}
\vspace{-0.5em}
\resizebox{1.0\linewidth}{!}{
\begin{tabular}{lcccccccccccc}
\toprule
Dataset      & \multicolumn{6}{c}{Mip-NeRF 360}                & \multicolumn{6}{c}{Tanks\&Temples}              \\\cmidrule(lr){2-7}\cmidrule(lr){8-13}
Method       & PSNR  & SSIM  & LPIPS & Train  & FPS  & Storage & PSNR  & SSIM  & LPIPS & Train  & FPS  & Storage \\\midrule
Plenoxels    & 23.08 & 0.626 & 0.463 & 25m 49s & 6.79 & 2.1 GB   & 21.08 & 0.719 & 0.379 & 25m 05s  & 13.0 & 2.3 GB   \\
INGP-base    & 25.30 & 0.671 & 0.371 & 05m 37s  & 11.7 & 13 MB    & 21.72 & 0.723 & 0.330 & 05m 26s  & 17.1 & 13 MB    \\
INGP-big     & 25.59 & 0.699 & 0.331 & 07m 30s  & 9.43 & 48 MB    & 21.92 & 0.745 & 0.305 & 06m 59s  & 14.4 & 48 MB    \\
Mip-NeRF 360 & 27.69 & 0.792 & 0.237 & 48h    & 0.06 & 8.6 MB   & 22.22 & 0.759 & 0.257 & 48h    & 0.14 & 8.6 MB   \\
3DGS         & 27.21 & 0.815 & 0.214 & 41m 33s & 134  & 734 MB   & 23.14 & 0.841 & 0.183 & 26m 54s & 154  & 411 MB   \\\midrule
3DGS*         & 27.46 & 0.812 & 0.222 & 24m 07s & 120  & 746 MB   & 23.71 & 0.845 & 0.178 & 13m 51s & 160  & 432 MB   \\
Ours         & 27.08 & 0.798 & 0.247 & 33m 06s & \textbf{128}  & \textbf{48.8 MB}  & 23.32 & 0.831 & 0.201 & 18m 20s & \textbf{185}  & \textbf{39.4 MB} \\
Ours+PP         & 27.03 & 0.797 & 0.247 & - & -  & \textbf{29.1 MB}  & 23.32 & 0.831 & 0.202 & - & -  & \textbf{20.9 MB}
\\\bottomrule
\end{tabular}}
\vspace{-0.5em}
\label{tab:qual1}
\end{table*}

\begin{table}[ht]
\caption{Qualitative results of the proposed method evaluated on Deep Blending dataset. We re-evaluate 3DGS* same with our method.}
\vspace{-0.5em}
\resizebox{1.0\linewidth}{!}{
\begin{tabular}{lcccccc}
\toprule
Dataset      & \multicolumn{6}{c}{Deep Blending}                \\\cmidrule(lr){2-7}
Method       & PSNR  & SSIM  & LPIPS & Train   & FPS  & Storage \\\midrule
Plenoxels    & 23.06 & 0.795 & 0.510 & 27m 49s & 11.2 & 2.7 GB  \\
INGP-base    & 23.62 & 0.797 & 0.423 & 06m 31s & 3.26 & 13 MB   \\
INGP-big     & 24.96 & 0.817 & 0.390 & 08m 00s & 2.79 & 48 MB   \\
Mip-NeRF 360 & 29.40 & 0.901 & 0.245 & 48h     & 0.09 & 8.6 MB  \\
3DGS         & 29.41 & 0.903 & 0.243 & 36m 02s & 137  & 676 MB  \\\midrule
3DGS*         & 29.46 & 0.900 & 0.247 & 21m 52s & 132  & 663 MB  \\
Ours         & \textbf{29.79} & \textbf{0.901} & 0.258 & 27m 33s & \textbf{181}  & \textbf{43.2MB}\\
Ours+PP         & 29.73 & 0.900 & 0.258 & - & -  & \textbf{23.8MB}\\\bottomrule
\end{tabular}}
\vspace{-1em}
\label{tab:qual2}
\end{table}

\subsection{Training}
Here, we have $N$ Gaussians and their attributes, position $p_n$, opacity $o_n$, rotation $\hat{r}_n$, scale $\hat{s}_n$, and view-dependent color $c_n(\cdot)$, which are used to render images. The entire model is trained end-to-end based on the rendering loss $L_{ren}$, the weighted sum of the L1 and SSIM loss between the GT and rendered images.
By adding the loss for masking $L_m$ and geometry codebooks $L_r, L_s$, the overall loss $L$ is as follows,
\begin{align}
    &L = L_{ren} + \lambda_m L_m + L_r + L_s,
\end{align}
where $\lambda_m$ is a hyperparameter to regularize the number of Gaussians.
To avoid heavy computations and ensure fast and optimal training, we apply R-VQ and learn codebooks with K-means initialization, only for the last 1K training iterations.
Except for that period, we set $L_r,L_s$ to zero.

\section{Experiment}

\subsection{Implementation Details}
We tested our approach on three real-world datasets (Mip-NeRF 360~\cite{mip360}, Tanks\&Temples~\cite{tnt}, and Deep Blending~\cite{db}) and a synthetic dataset (NeRF-Synthetic~\cite{nerf}).
Following 3DGS, we chose two scenes from Tanks\&Temples and Deep Blending.
The models with the proposed method (denoted as Ours) were trained during 30K iterations, where we stored positions $p$ and opacities $o$ with 16-bit precision using half-tensors.
Additionally, we implemented straightforward post-processing techniques on the model attributes, a variant we denote as Ours+PP. These post-processing steps include:
\begin{itemize}
    \item Applying 8-bit min-max quantization to opacity and hash grid parameters.
    \item Pruning hash grid parameters with values below 0.1.
    \item Applying Huffman encoding~\cite{huffman} on the quantized opacity and hash parameters, and R-VQ indices.
\end{itemize}
Further implementation details are provided in the supplementary materials.


\begin{table}[t]
\centering
\caption{Qualitative results of the proposed method evaluated on NeRF-Synthetic dataset. * denotes the reported value in the original paper.}
\vspace{-0.5em}
\resizebox{1.0\linewidth}{!}{
\begin{tabular}{lcccc}\toprule
Dataset & \multicolumn{4}{c}{NeRF-Synthetic}                     \\
\cmidrule(lr){2-5}
Method  & PSNR  & Train Time & Storage FPS & FPS \\\midrule
3DGS    & 33.32* &  68.1 MB                        & 6m 14s & 359 \\
Ours    & 33.33 & \textbf{5.55 MB} ($\times$0.08)                         & 8m 04s ($\times$1.29)  & \textbf{545} ($\times$1.52)\\
Ours+PP    & 32.88 & \textbf{2.67 MB} ($\times$0.04)                         & -  & - \\\bottomrule
\end{tabular}}
\vspace{-1em}
\label{tab:qual3}
\end{table}

\begin{figure*}[ht]
    \begin{center}
    \includegraphics[width=1.0\linewidth]{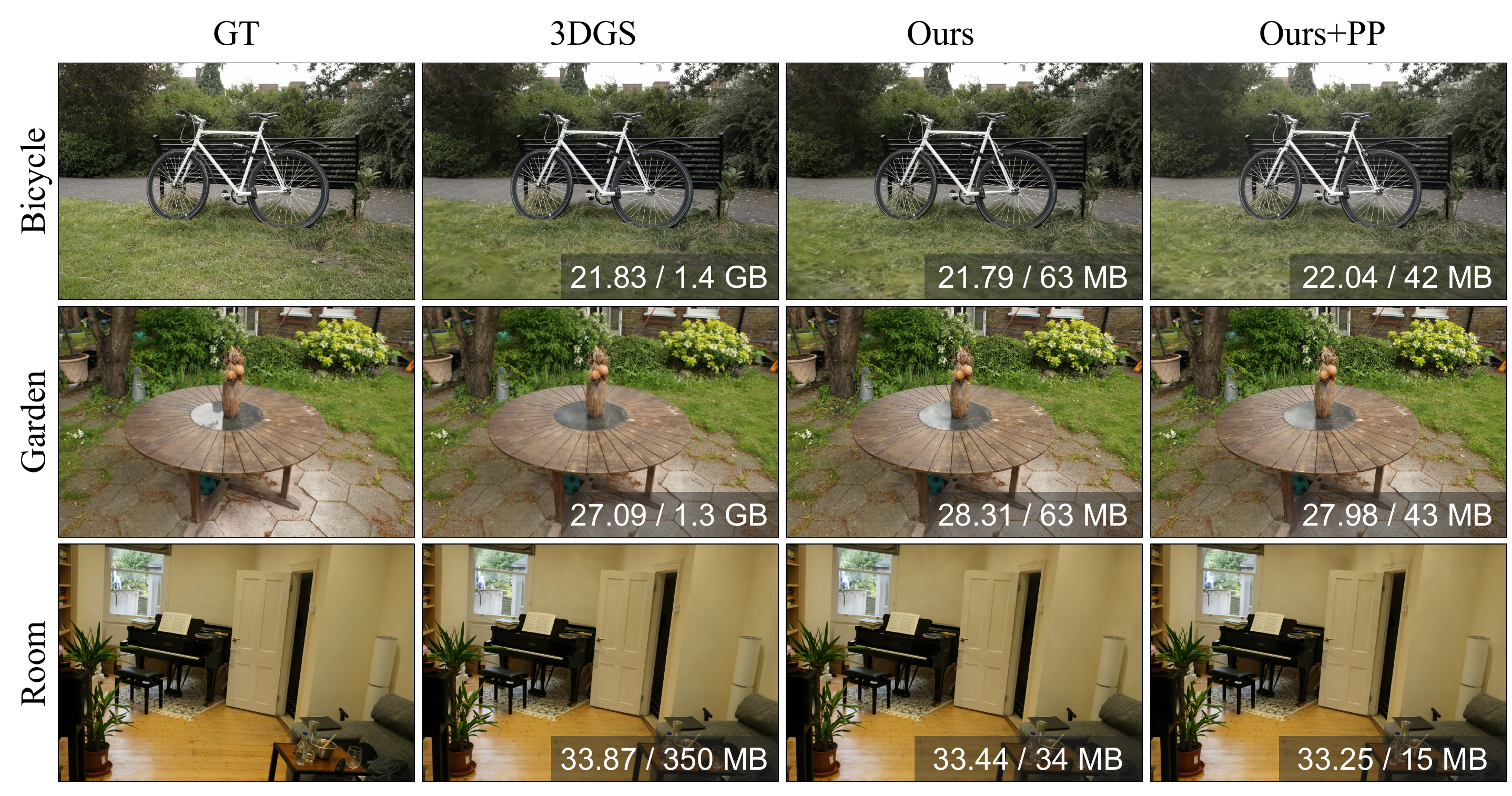}
    \end{center}
    \vspace{-1.5em}
    \caption{Qualitative results of our approach compared to 3DGS. We present the rendering PSNR and storage on the results.}
\label{fig:qual}
\end{figure*}

\begin{figure*}[ht]
    \begin{center}
    \includegraphics[width=1.0\linewidth]{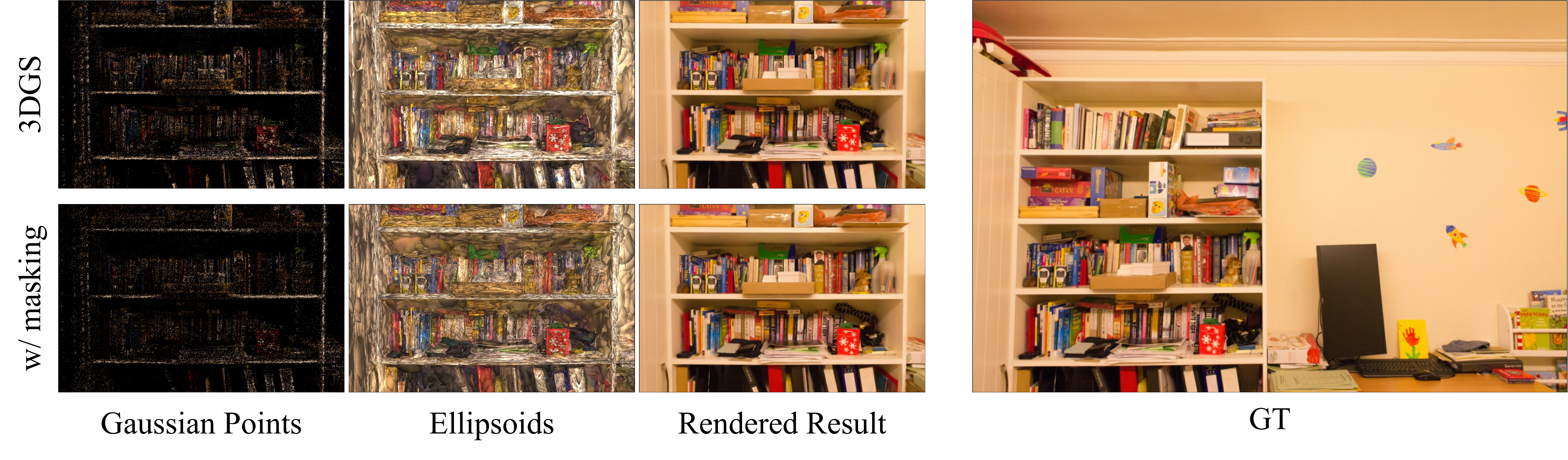}
    \end{center}
    \vspace{-1.5em}
    \caption{Effect of the proposed learnable volume masking, compared to the original 3DGS. We visualize Gaussian center points, ellipsoids, and rendered results using \textit{Playroom} scene.}
\label{fig:vis_mask}
\end{figure*}

\begin{figure*}[ht]
    \begin{center}
    \includegraphics[width=1.0\linewidth]{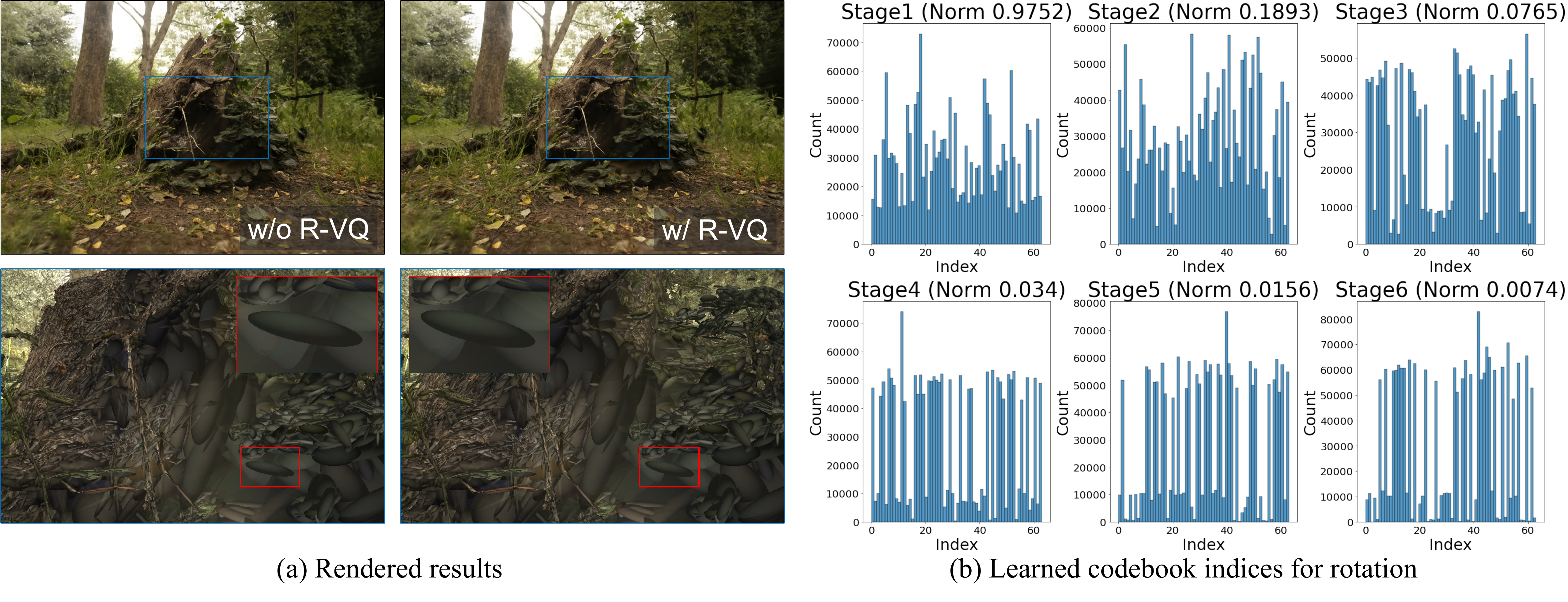}
    \end{center}
    \vspace{-1.5em}
    \caption{Effect of the proposed geometry codebook. We visualize (a) ellipsoids and rendered results and (b) learned codebook indices for rotation using \textit{Stump} scene. `Norm' denotes the average norm of all code vectors in each codebook (representing magnitude).}
\label{fig:vis_rvq}
\end{figure*}

\begin{table*}[ht]
\centering
\caption{Ablation study on the proposed contributions. `M', `C', `G', and `H' denote masking, color representation, geometry codebook, and half tensor for positions and opacities, respectively. `\#Gauss' means the number of Gaussians.}
\vspace{-0.5em}
\begin{tabular}{cccccccccccccc}\toprule
\multicolumn{4}{c}{Method \textbackslash Dataset} & \multicolumn{5}{c}{Playroom}                 & \multicolumn{5}{c}{Bonsai}                   \\
\cmidrule(lr){1-4}\cmidrule(lr){5-9}\cmidrule(lr){10-14}
M       & C       & G       & H      & PSNR  & Train time & \#Gauss & Storage & FPS & PSNR  & Train time & \#Gauss & Storage & FPS \\\midrule
\multicolumn{4}{c}{3DGS}                          & 29.87 & 19m 22s    & 2.34 M  & 553 MB  & 154 & 32.16 & 19m 18s    & 1.25 M  & 295 MB  & 200 \\
\checkmark          &             &            &           & 29.91 & 17m 51s    & 967 K   & 228 MB  & 254 & 32.22 & 18m 50s    & 643 K   & 152 MB  & 247 \\
\checkmark          & \checkmark          &            &           & 30.33 & 23m 56s    & 770 K   & 59 MB   & 210 & 32.08 & 23m 09s    & 592 K   & 51 MB   & 196 \\
\checkmark          & \checkmark          & \checkmark         &           & 30.33 & 24m 58s    & 761 K   & 44 MB   & 204 & 32.08 & 24m 06s    & 598 K   & 40 MB   & 198 \\
\checkmark        & \checkmark         & \checkmark        & \checkmark       & 30.32 & 24m 35s    & 778 K   & 38 MB   & 206 & 32.08 & 24m 16s    & 601 K   & 35 MB   & 196 \\\bottomrule
\end{tabular}
\vspace{-1em}
\label{tab:abl}
\end{table*}

\subsection{Performance Evaluation}
\noindent\textbf{Real-world scenes.} \cref{tab:qual1} and \cref{tab:qual2} show the qualitative results evaluated on real-world scenes.
Our approach consistently reduces the storage requirements significantly and enhances the rendering speed, while performing accurate reconstruction comparable to 3DGS.
Especially for the Deep Blending dataset (\cref{tab:qual2}), our method even outperforms the original 3DGS in terms of visual quality (measured in PSNR and SSIM), achieving state-of-the-art performance with the fastest rendering speed as well as compactness (almost 40\% faster rendering and over 15$\times$ compactness compared to 3DGS).
The qualitative results in \cref{fig:qual} also show the high-quality reconstruction with significantly reduced storage of our method, compared to 3DGS. 

\noindent\textbf{Synthetic scenes.} We also evaluate our method on synthetic scenes.
As 3DGS has proven its effectiveness in visual quality, rendering speed, and training time compared to other baselines, we focus on comparing our method with 3DGS, highlighting the improvements from our method. As shown in \cref{tab:qual3}, although our approach requires slightly more training duration, we achieve over 10$\times$ compression and 50\% faster rendering compared to 3DGS, maintaining high-quality reconstruction.

\noindent\textbf{With post-processings,} our model is further downsized by over 40 \% regardless of dataset. 
Consequently, we achieve more than 25$\times$ compression from 3DGS, while maintaining high performance.

\subsection{Ablation Study}

\noindent\textbf{Learnable volume masking.}
As shown in \cref{tab:abl}, the proposed volume-based masking reduces the number of Gaussians significantly while retaining (even slightly increasing) the visual quality, demonstrating its effective removal of redundant and unessential Gaussians.
The reduced Gaussians show several additional advantages: reducing training time, storage, and testing time.
Specifically on \textit{Playroom} scene, the volume-mask shows a 140\% increase in storage efficiency and a 65\% increase in rendering speed.
Furthermore, we validate the actual impact in rendering by visualizing Gaussians in \cref{fig:vis_mask}.
Although the number of Gaussians is noticeably reduced demonstrated by the sparser points in visualization, the rendered result retains high quality without visible difference.
These results quantitatively and qualitatively demonstrate the effectiveness and efficiency of the proposed method.

\noindent\textbf{Compact view-dependent color.}
The proposed color representation based on the neural field offers more than a threefold improvement in storage efficiency with a slightly reduced number of Gaussians compared to the directly storing high-degree SH, despite necessitating slightly more time for training and rendering. 
Nonetheless, when compared to 3DGS, the proposed color representation with the masking strategy demonstrates either a faster or comparable rendering speed.

\noindent\textbf{Geometry codebook.}
Our proposed geometry codebook approach achieves a reduction in storage requirements by approximately 30\% while maintaining the quality of reconstruction, training time, and rendering speed. To delve deeper into the reasons for this improved performance, we start by visualizing the Gaussians, as depicted in \cref{fig:vis_rvq}-(a). It is observed that the majority of Gaussians maintain identical geometry in the visualizations regardless of the application of R-VQ, with only a few exhibiting very subtle geometric distortions that are hardly noticeable.
Although these results prove the effectiveness of our method, we also explore the patterns of learned indices across each stage of R-VQ, shown in \cref{fig:vis_rvq}-(b). The initial stage exhibits an even distribution with a large magnitude of codes.
As the stages progress, the magnitude of codes decreases, indicating that the residuals of each stage have been effectively trained to represent geometry.

\begin{table}[ht]
\caption{Avarge storage (MB) for each Gaussian attribute, evaluated on Mip-NeRF 360 dataset. f, 8b, H, and P mean floating-point, min-max quantization to 8-bit, Huffman encoding, and pruning parameters below 0.1, respectively.}
\vspace{-0.3em}
\resizebox{1.0\linewidth}{!}{
\begin{tabular}{lccccccc}
    \toprule
     & Pos. & Opa. & Sca. & Rot. & \multicolumn{2}{c}{Col.} & Tot.  \\\midrule
    \multirow{2}{*}{3DGS} & \multicolumn{6}{c}{32f} & \multirow{2}{*}{746}  \\\cmidrule(lr){2-7}
     & 37.9  & 12.6  & 37.9  & 50.6  & \multicolumn{2}{c}{606.9} &   \\\midrule
    \multirow{2}{*}{Ours} & \multicolumn{2}{c}{16f} & \multicolumn{2}{c}{R-VQ} & Hash(16f) & MLP(16f) & \multirow{2}{*}{48.8}  \\\cmidrule(lr){2-3}\cmidrule(lr){4-5}\cmidrule(lr){6-6}\cmidrule(lr){7-7}
     & 8.3  & 2.8  & 6.3  & 6.3  & 25.2 & 0.016  &   \\\midrule
    \multirow{2}{*}{\begin{tabular}[l]{@{}l@{}}Ours\\+PP\end{tabular}} & 16f & 8b+H & \multicolumn{2}{c}{+H} & +8b+P+H & MLP(16f) & \multirow{2}{*}{29.1} \\\cmidrule(lr){2-2}\cmidrule(lr){3-3}\cmidrule(lr){4-5}\cmidrule(lr){6-6}\cmidrule(lr){7-7}
     & 8.3  & 1.2  & 5.9  & 6.2  & 7.4  & 0.016  & 
    \\\bottomrule
    \end{tabular}}
\label{tab:size}
\end{table}

\noindent\textbf{Effect of post-processing. }
\cref{tab:size} describes the size of each attribute with and without the application of post-processing techniques. While our end-to-end trainable framework demonstrates significant effectiveness, it requires relatively large storage for color representation. 
Nonetheless, as indicated in the table, this can be effectively reduced through simple post-processings.
    
\vspace{-0.2em}
\section{Conclusion}
We have proposed a compact 3D Gaussian representation for 3D scenes, reducing the number of Gaussian points without performance decrease through a novel volume-based masking.
Furthermore, this work proposed combining the neural field and exploiting the learnable codebooks to represent Gaussian attributes compactly.
In the extensive experiments, our approach demonstrated more than a tenfold reduction in storage and a marked increase in rendering speed while retaining the high-quality reconstruction compared to 3DGS.
This result sets a new benchmark with high performance, fast training, compactness, and real-time rendering. 
Our framework thus stands as a comprehensive solution, paving the way for broader adoption and application in various fields requiring efficient and high-quality 3D scene representation.
{
    \small
    \bibliographystyle{ieeenat_fullname}
    \bibliography{main}
}

\newpage
\section*{Appendix}

\section{Implementation Detail}
We retained all hyper-parameters of 3DGS and trained models during 30K iterations, and we set the codebook size $C$ and the number of stages $L$ of R-VQ to 64 and 6, respectively.
The neural field for view-dependent color uses hash grids with 2-channel features across 16 different resolutions (16 to 4096) and a following 2-layer 64-channel MLP.
Due to the different characteristics between the real and synthetic scenes, we adjusted the maximum hash map size and the hyper-parameters for learning the neural field and the mask.
For the real scenes, we set the max size of hash maps to $2^{19}$, the control factor for the number of Gaussians $\lambda_m$ to $5e^{-4}$, and the learning rate of the mask parameter and the neural fields to $1e^{-2}$.
The learning rate of the neural fields is decreased at 5K, 15K, and 25K iterations by multiplying a factor of $0.33$.
For the synthetic scenes, the maximum hash map size and the control factor $\lambda_m$ were set to $2^{16}$ and $4e^{-3}$, respectively.
The learning rate of the mask parameter and the neural fields were set to $1e^{-3}$, where the learning rate of the neural fields was reduced at 25K iteration with a factor of $0.33$.

\section{Fast inference pipeline}
The main paper's analysis shows that our method extends the overall training time slightly more than 3DGS, owing to the time for iterating neural fields and for R-VQ search. However, our approach effectively reduces rendering time for several reasons. First, the proposed masking strategy significantly effectively reduces the number of Gaussians, as demonstrated in our results, leading to reduced training and rendering times. Second, by precomputing grid features at the testing phase, we minimize the operational time of the neural field. Since these grid features, which precede the subsequent MLP input, are not dependent on the view direction itself, they can be prepared in advance of testing. This allows our method to simply process a small MLP for generating view-dependent colors during testing. Third, in the testing phase, the time spent searching for suitable geometry is eliminated. In a manner similar to precomputing grid features, we can index the closest codes from multi-stage codebooks before testing. These strategies collectively enable us to achieve a notably faster rendering speed.

\begin{figure*}[t]
    \centering
    \includegraphics[width=1.0\linewidth]{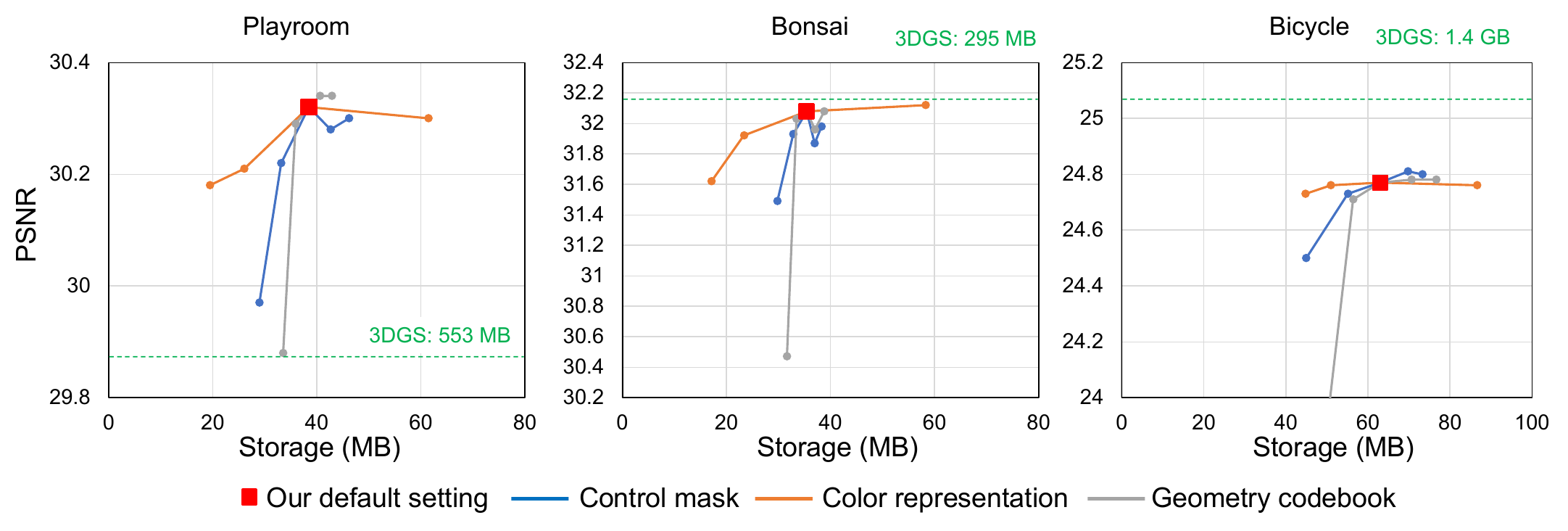}
    \caption{Rate-distortion curves evaluated on diverse scenes. Starting with our default model, we methodically adjust the compactness level of each proposal to evaluate their individual contributions.}
\label{fig:app_abl}
\end{figure*}

\section{Additional ablation study}
\subsection{Rate-distortion curve based on each proposal}
In addition to the ablation study in the main paper, we conduct an in-depth analysis on the impact of our contributions: volume-based masking, compact color representation, and geometry codebook.
As illustrated in \cref{fig:app_abl}, we start with the standard configuration of our approach and adjust the compactness level of the three proposals. We control $\lambda_m$, max hashmap size, and the number of R-VQ stages, respectively, by doubling each hyper-parameter.
As the performance for the different scenes varies due to their diverse characteristics, we choose three distinct scenes where our approach increases, retains, and decreases visual quality while achieving over 10$\times$ compression.

Although the proposed geometry codebook effectively reduces the storage as validated in the main paper, it shows poor performance when the diversity of codebooks is extremely limited.
In contrast, our neural field-based color representation demonstrates remarkable robustness even in a low-rate condition across the various scenes, indicating that scaling down the neural field is the best option in environments with severe resource limitations.

These results show that the storage need of our method can be halved with minimal performance loss, and our default configuration is a well-rounded choice for a wide range of scenes.

\begin{table}[h]
\centering
\begin{tabular}{ccccc}
\toprule
Opacity            & Scale              & $\lambda_m$       & \#Gaussian      & PSNR           \\\midrule
\multirow{2}{*}{\checkmark} & \multirow{2}{*}{}  & 0.0005          & 311786          & 31.50          \\
                   &                    & 0.0001          & 629837          & 31.86          \\\cmidrule(lr){1-3}\cmidrule(lr){4-5}
\multirow{2}{*}{}  & \multirow{2}{*}{\checkmark} & 0.0005          & 508762          & 31.89          \\
                   &                    & 0.0003          & 596449          & 31.97          \\\cmidrule(lr){1-3}\cmidrule(lr){4-5}
\textbf{\checkmark}         & \textbf{\checkmark}         & \textbf{0.0005} & \textbf{601048} & \textbf{32.08}
\\\bottomrule
\end{tabular}
\caption{Ablation study on the proposed volume-based masking. \#Gaussian denotes the number of Gaussians}
\label{abl_volmask}
\end{table}

\begin{table}[]
\centering
\caption{Ablation study on represented components by I-NGP.}
\label{tab:abl_ingp}
\resizebox{1.0\linewidth}{!}{
\begin{tabular}{ccccccc}
\toprule
\multicolumn{4}{c}{I-NGP representation} & \multicolumn{1}{l}{\multirow{2}{*}{PSNR}} & \multicolumn{1}{l}{\multirow{2}{*}{Train time}} & \multicolumn{1}{l}{\multirow{2}{*}{\#Gaussians}} \\
Opa. & Sca. & Rot. & Col. & \multicolumn{1}{l}{} & \multicolumn{1}{l}{} & \multicolumn{1}{l}{} \\\midrule
\multicolumn{4}{c}{3DGS} & 32.2 & 19:21 & 1245 K \\\cmidrule(lr){1-4}\cmidrule(lr){5-7}
 &  &  & \checkmark & 32.3 & 24:42 & 1178 K \\\cmidrule(lr){1-4}\cmidrule(lr){5-7}
\checkmark &  &  & \checkmark & 9.3 & 29:55 & 666 K \\
 & \checkmark &  & \checkmark & 9.3 & 29:55 & 559 K \\
 &  & \checkmark & \checkmark & 25.9 & 27.37 & 1692 K
\\\bottomrule
\end{tabular}}
\end{table}

\begin{table*}[ht]
\centering
\resizebox{1.0\textwidth}{!}{
\begin{tabular}{lcccccccccccc}
\toprule
Scene  & \multicolumn{3}{c}{bicycle} & \multicolumn{3}{c}{bonsai} & \multicolumn{3}{c}{drjohnson} & \multicolumn{3}{c}{playroom} \\\cmidrule(lr){1-1}\cmidrule(lr){2-4}\cmidrule(lr){5-7}\cmidrule(lr){8-10}\cmidrule(lr){11-13}
Method & PSNR   & Storage  & Mem.    & PSNR   & Storage  & Mem.   & PSNR    & Storage   & Mem.    & PSNR    & Storage  & Mem.    \\\cmidrule(lr){1-1}\cmidrule(lr){2-4}\cmidrule(lr){5-7}\cmidrule(lr){8-10}\cmidrule(lr){11-13}
3DGS   & 25.08  & 1.4 GB   & 9.4 GB  & 32.16  & 295 MB   & 8.7 GB & 29.06   & 774 MB    & 7.5 GB  & 29.87   & 553 MB   & 6.4 GB  \\
Ours   & 24.77  & 63 MB    & 7.6 GB  & 32.08  & 35 MB    & 8.3 GB & 29.26   & 48 MB     & 6.5 GB  & 30.32   & 39 MB    & 5.6 GB 
\\\bottomrule
\end{tabular}}
\caption{Evaluation of GPU memory requirements for our method compared to 3DGS. 'Mem' indicates the GPU memory requirements.}
\label{tab:infmem}
\end{table*}

\subsection{Volume-based masking}
We have proposed the learnable masking of Gaussians based on their volume as well as transparency.
As shown in \cref{abl_volmask}, masking based on only Gaussian volume outperforms the method that only considers Gaussian opacity.
Moreover, the best results are achieved when both volume and transparency are taken into account for masking.

\subsection{Usage of I-NGP and R-VQ.}
Neural fields can represent continuous signals efficiently, whereas VQ works well for repetitive components. 
In a 3D scene, near Gaussians can be expected to share similar color but are not guaranteed to have a similar shape, rather, similar shapes can be frequently found in the entire scene.
Therefore, I-NGP is not successfully trained to represent other attributes except for color, as demonstrated in \cref{tab:abl_ingp}.
Given this intuition, we have proposed an effective representation for each component.

\section{Inference memory}
Throughout the main paper, we analyze the effectiveness of our method in terms of the balance between the visual quality, storage requirement, and rendering speed. Our approach effectively reduces not only storage needs but also memory requirements, both of which are key aspects of efficiency.
\cref{tab:infmem} depicts the GPU memory requirement for inference with the visual quality and storage need, evaluated on diverse scenes.
Our method consistently reduces the GPU memory requirements by a safe margin, demonstrating its efficient usage of computing resources.

\section{Per-Scene Results}
We evaluated the performance on various datasets for novel view synthesis.
We provide per-scene results for Mip-NeRF 360 (\cref{tab:per_360}), Tanks\&Temples (\cref{tab:per_td}), Deep Blending (\cref{tab:per_td}), and NeRF synthetic (\cref{tab:per_nerf}) datasets.

\begin{table*}[]
\resizebox{1.0\textwidth}{!}{
\begin{tabular}{clcccccccccc}
\toprule
\multicolumn{2}{c}{Scene}                     & bicycle & flowers & garden  & stump   & tree hill & room    & counter & kitchen & bonsai  & Avg.    \\\midrule
\multirow{7}{*}{3DGS} & PSNR & 25.10   & 21.33   & 27.25   & 26.66   & 22.53    & 31.50   & 29.11   & 31.53   & 32.16   & 27.46   \\
                      &           SSIM            & 0.747   & 0.588   & 0.856   & 0.769   & 0.635    & 0.925   & 0.914   & 0.932   & 0.946   & 0.812   \\
                      & LPIPS                 & 0.244   & 0.361   & 0.122   & 0.243   & 0.346    & 0.198   & 0.184   & 0.117   & 0.181   & 0.222   \\
                      & Train (mm:ss)         & 34:04   & 25:33   & 33:46   & 27:05   & 24:51    & 22:55   & 22:42   & 26:08   & 19:18   & 24:07   \\
                      & \#Gaussians           & 5,723,640 & 3,414,994 & 5,641,235 & 4,549,202 & 3,470,681  & 1,483,653 & 1,171,684 & 1,744,761 & 1,250,329 & 3,161,131 \\
                      & Storage (MB)          & 1350.78 & 805.94  & 1331.33 & 1073.60 & 819.08   & 350.14  & 276.52  & 411.76  & 295.08  & 746.03  \\
                      & FPS                   & 63.81   & 132.03  & 77.19   & 108.81  & 100.92   & 132.51  & 146.40  & 122.07  & 199.86  & 120.40  \\\cmidrule(lr){1-2}\cmidrule(lr){3-11}\cmidrule(lr){12-12}
\multirow{7}{*}{Ours} & PSNR                  & 24.77   & 20.89   & 26.81   & 26.46   & 22.65    & 30.88   & 28.71   & 30.48   & 32.08   & 27.08   \\
                      & SSIM                  & 0.723   & 0.556   & 0.832   & 0.757   & 0.638    & 0.919   & 0.902   & 0.919   & 0.939   & 0.798   \\
                      & LPIPS                 & 0.286   & 0.399   & 0.161   & 0.278   & 0.363    & 0.209   & 0.205   & 0.131   & 0.193   & 0.247   \\
                      & Train (mm:ss)         & 42:36   & 32:37   & 45:36   & 33:43   & 34:08    & 24:18   & 27:41   & 32:59   & 24:16   & 33:06   \\
                      & \#Gaussians           & 2,221,689 & 1,525,598 & 2,209,609 & 1,732,089 & 2,006,446  & 529,136  & 536,672  & 1,131,168 & 601,048  & 1,388,162 \\
                      & Storage (MB)          & 62.99   & 51.15   & 62.78   & 54.66   & 59.33    & 34.21   & 34.34   & 44.45   & 35.44   & \textbf{48.82}   \\
                      & FPS                   & 76.41   & 142.41  & 89.49   & 120.96  & 110.28   & 183.03  & 119.52  & 114.24  & 196.08  & \textbf{128.05} \\\cmidrule(lr){1-2}\cmidrule(lr){3-11}\cmidrule(lr){12-12}
\multirow{4}{*}{Ours+PP} & PSNR                  & 24.73   & 20.89   & 26.72   & 26.31   & 22.67    & 30.88   & 28.63   & 30.48   & 31.98   & 27.03   \\
                      & SSIM                  & 0.722   & 0.554   & 0.831   & 0.754   & 0.637    & 0.918   & 0.901   & 0.919   & 0.937   & 0.797   \\
                      & LPIPS                 & 0.284   & 0.399   & 0.158   & 0.280   & 0.363    & 0.209   & 0.206   & 0.130   & 0.193   & 0.247   \\
                      & Storage (MB)          & 42.42   & 32.05   & 43.26   & 33.83   & 39.08    & 15.01   & 15.22   & 24.39   & 16.40   & \textbf{29.07} 
\\\bottomrule
\end{tabular}}
\caption{Per-scene results evaluated on Mip-NeRF 360 dataset.}
\label{tab:per_360}
\end{table*}

\begin{table*}[]
\centering
\begin{tabular}{clcccccc}\toprule
\multicolumn{2}{c}{Dataset}                  & \multicolumn{3}{c}{Tanks\&Temples}                                         & \multicolumn{3}{c}{Deep Blending}      \\\cmidrule(lr){1-2}\cmidrule(lr){3-5}\cmidrule(lr){6-8}
\multicolumn{2}{c}{Scene}                    & train                      & truck                      & Avg.             & drjohnson & playroom & Avg.            \\\midrule
\multirow{7}{*}{3DGS} & PSNR                 & 22.07                      & 25.35                      & 23.71            & 29.06     & 29.87    & 29.46           \\
                      & SSIM                 & 0.812                      & 0.878                      & 0.845            & 0.899     & 0.901    & 0.900           \\
                      & LPIPS                & 0.208                      & 0.148                      & 0.178            & 0.247     & 0.247    & 0.247           \\
                      & Train (mm:ss)        & 12:18                      & 15:24                      & 13:51            & 24:22     & 19:22    & 21:52           \\
                      & \#Gaussians & 1,084,001           & 2,579,252           & 1,831,627 & 3,278,027   & 2,343,368  & 2,810,698         \\
                      & Storage (MB)         & 255.82                     & 608.70                     & 432.26           & 773.61    & 553.03   & 663.32          \\
                      & FPS                  & 174.42 & 145.14 & 159.78           & 110.46    & 154.47   & 132.47          \\\cmidrule(lr){1-2}\cmidrule(lr){3-5}\cmidrule(lr){6-8}
\multirow{7}{*}{Ours} & PSNR                 & 21.56                      & 25.07                      & 23.32            & 29.26     & 30.32    & \textbf{29.79}  \\
                      & SSIM                 & 0.792                      & 0.871                      & 0.831            & 0.900     & 0.902    & 0.901           \\
                      & LPIPS                & 0.240                      & 0.163                      & 0.201            & 0.258     & 0.258    & 0.258           \\
                      & Train (mm:ss)        & 16:03                    & 20:36                    & 18:20            & 30:31   & 24:35  & 27:33        \\
                      & \#Gaussians          & 710,434                     & 962,158                     & 836,296           & 1,339,005   & 778,353   & 1,058,679         \\
                      & Storage (MB)         & 37.29                      & 41.57                      & \textbf{39.43}   & 47.98     & 38.45    & \textbf{43.21}  \\
                      & FPS                  & 185.91                     & 184.83                     & \textbf{185.37}  & 155.37    & 205.83   & \textbf{180.60}
                      \\\cmidrule(lr){1-2}\cmidrule(lr){3-5}\cmidrule(lr){6-8}
\multirow{4}{*}{Ours+PP} & PSNR                 & 21.62                      & 25.02                      & 23.32            & 29.16     & 30.30    & \textbf{29.73}  \\
                      & SSIM                 & 0.792                      & 0.870                      & 0.831            & 0.899     & 0.900    & 0.900           \\
                      & LPIPS                & 0.240                      & 0.163                      & 0.202            & 0.257     & 0.259    & 0.258           \\
                      & Storage (MB)         & 19.07                      & 22.64                      & \textbf{20.86}   & 28.43     & 19.21    & \textbf{23.82}  \\\bottomrule
\end{tabular}
\caption{Per-scene results evaluated on Tank\&Temples and Deep Blending.}
\label{tab:per_td}
\end{table*}

\begin{table*}[]
\centering
\begin{tabular}{clccccccccc}
\toprule
\multicolumn{2}{c}{Scene} & chair & drums & ficus & hotdog & lego & materials & mic & ship & Avg. \\\cmidrule(lr){1-2}\cmidrule(lr){3-10}\cmidrule(lr){11-11}
\multirow{7}{*}{Ours} & PSNR & 34.91 & 26.18 & 35.44 & 37.38 & 35.48 & 29.97 & 35.81 & 31.51 & 33.33 \\
 & SSIM & 0.986 & 0.953 & 0.987 & 0.984 & 0.981 & 0.958 & 0.991 & 0.905 & 0.968 \\
 & LPIPS & 0.013 & 0.041 & 0.013 & 0.023 & 0.018 & 0.042 & 0.008 & 0.113 & 0.034 \\
 & Train (m:ss) & 9:19 & 8:55 & 6:41 & 8:20 & 8:23 & 6:53 & 7:12 & 8:46 & 8:04 \\
 & \#Gaussians & 153,570 & 178,615 & 83,910 & 64,194 & 171,826 & 107,188 & 56,015 & 148,442 & 120,470 \\
 & Storage (MB) & 6.11 & 6.54 & 4.93 & 4.59 & 6.42 & 5.32 & 4.44 & 6.02 & \textbf{5.55} \\
 & FPS & 512.10 & 427.56 & 706.71 & 719.85 & 402.15 & 638.01 & 674.34 & 282.72 & 545.43 \\
 \cmidrule(lr){1-2}\cmidrule(lr){3-10}\cmidrule(lr){11-11}
\multirow{4}{*}{Ours+PP} & PSNR & 34.58 & 26.01 & 35.06 & 36.71 & 34.96 & 29.04 & 35.57 & 31.06 & 32.88 \\
 & SSIM & 0.985 & 0.951 & 0.987 & 0.983 & 0.979 & 0.954 & 0.991 & 0.903 & 0.968 \\
 & LPIPS & 0.013 & 0.042 & 0.013 & 0.023 & 0.020 & 0.043 & 0.008 & 0.115 & 0.034 \\
 & Storage (MB) & 3.09 & 3.56 & 2.11 & 1.75 & 3.45 & 2.51 & 1.59 & 3.28 & \textbf{2.67} \\\bottomrule
\end{tabular}
\caption{Per-scene results evaluated on NeRF-synthetic dataset.}
\label{tab:per_nerf}
\end{table*}

\end{document}